\documentclass{article}

\usepackage[final,nonatbib]{nips_2016}

\usepackage[utf8]{inputenc} % allow utf-8 input
\usepackage[T1]{fontenc}    % use 8-bit T1 fonts
\usepackage{hyperref}       % hyperlinks
\usepackage{url}            % simple URL typesetting
\usepackage{booktabs}       % professional-quality tables
\usepackage{amsfonts}       % blackboard math symbols
\usepackage{nicefrac}       % compact symbols for 1/2, etc.
\usepackage{microtype}      % microtypography
\usepackage{amsmath}
\usepackage{multirow}
\usepackage{placeins}
\usepackage{algorithm}
\usepackage{algpseudocode}
\usepackage{multicol}

\usepackage{url}
\usepackage{graphicx}
\usepackage{color}

\DeclareGraphicsExtensions{.png,.pdf}
\graphicspath{{imglocal/}{img/}{./}}
\usepackage{caption}
\usepackage{subcaption}
\usepackage{amsthm}
\usepackage{amsfonts}

\title{Supervised learning based on temporal coding in spiking neural networks}

\author{Hesham Mostafa \\
Department of Bioengineering, Jacobs School of Engineering \\
Institute of Neural Computation\\
UC San Diego, La Jolla, CA 92093 USA \\
Email: hmmostafa@ucsd.edu\\
}

\begin{document}
\maketitle

%% % \nipsfinalcopy is no longer used

\begin{abstract}
  Gradient descent training techniques are remarkably successful in training analog-valued artificial neural networks (ANNs). Such training techniques, however, do not transfer easily to spiking networks due to the spike generation hard non-linearity and the discrete nature of spike communication. We show that in a feedforward spiking network that uses a temporal coding scheme where information is encoded in spike times instead of spike rates, the network input-output relation is differentiable almost everywhere. Moreover, this relation is piece-wise linear after a transformation of variables. Methods for training ANNs thus carry directly to the training of such spiking networks as we show when training on the permutation invariant MNIST task. In contrast to rate-based spiking networks that are often used to approximate the behavior of ANNs, the networks we present spike much more sparsely and their behavior can not be directly approximated by conventional ANNs. Our results highlight a new approach for controlling the behavior of spiking networks with realistic temporal dynamics, opening up the potential for using these networks to process spike patterns with complex temporal information.
\end{abstract}


% \fi

\section{Introduction}
Artificial neural networks (ANNs) are enjoying great success as a means of learning complex non-linear transformations by example~\cite{LeCun_etal15}. The idea of a distributed network of simple neuron elements that adaptively adjusts its connection weights based on training examples is partially inspired by the operation of biological spiking networks~\cite{Rumelhart_McClelland86}. ANNs, however, are fundamentally different from spiking networks. Unlike ANN neurons that are analog-valued, spiking neurons communicate using all-or-nothing discrete spikes. A spike triggers a trace of synaptic current in the target neuron. The target neuron integrates synaptic current over time until a threshold is reached, and then emits a spike and resets. Spiking networks are dynamical systems in which time plays a crucial role, while time is abstracted away in conventional feedforward ANNs.

%% Unlike ANNs, there is no general technique for training feeforward spiking neural networks. The encoding 
%% While several approaches have investigated training spiking neural networks in order to produce precise  

%% The success of ANNs trained using back-propagation has motivated many approaches that attempt to train multi-layer spiking networks using similar techniques~\cite{OConnor_etal13,Diehl_etal15,Cao_etal15} . The most immediate proble
%% for training spiking networks first train conventional feedforward ANNs and then translate the trained weights to spiking networks developed to approximate the behavior of the original ANNs~\cite{OConnor_etal13,Diehl_etal15,Cao_etal15}. The spiking networks obtained using these methods use rate coding where the spiking rate of a neuron encodes an analog quantity corresponding to the analog output of an ANN neuron. High spike rates would then mask the discrete nature of the spiking activity.    %A spiking network can approximate an ANN if the spike rate of a neuron  its analog value. 

%% Learning in spiking networks is complicated by the discrete all-or-nothing nature of neuronal spikes. 
%% Several previous approaches have been proposed for  
%% Moreover, our formulation results in an analytical relation between input and output spike times. This renders our approach much more computationally tractable than approaches    

ANNs typically make use of gradient descent techniques to solve the weight credit assignment problem~\cite{Minsky61}, that is the problem of changing the network weights so as to obtain the desired network output. ANNs typically have multiple cascaded layers of neurons. In that case, the gradient of the error function with respect to the network weights can be efficiently obtained using the backpropagation algorithm. Having multiple layers of neurons is crucial in allowing ANNs to learn using backpropagation~\cite{Ba_Caruna14}.

While backpropagation is a well-developed general technique for training feedforward ANNs, there is no general technique for training feedforward spiking neural networks. Many previous approaches that train spiking neural networks to produce particular spike patterns depend on having the input layer directly connected to the output layer~\cite{Memmesheimer_etal14,Gutig_Sompolinsky06}. It is unclear how multi-layer networks can be trained using these approaches. Stochastic network formulations are often considered when training temporal networks~\cite{Pfister_etal06,Gardner_etal15}. In a stochastic formulation, the goal is to maximize the likelihood of an entire output spike pattern. The stochastic formulation is needed to 'smear out' the discrete nature of the spike, and to work instead with spike generation probabilities that depend smoothly on network parameters and are thus more suitable for gradient descent learning. While multi-layer networks have been trained using this stochastic approach~\cite{Gardner_etal15}, there are several scalability concerns due to the need for time-stepped simulations to get output spike times. In some cases, Monte Carlo simulations are needed to obtain the likelihoods of different output patterns~\cite{Pfister_etal06}. Moreover, in previous approaches, the goal is to learn particular spike patterns, and the performance of these networks in classification settings where the goal is to learn an input-output relation that generalizes well to unseen examples, which are not merely noise corrupted training examples, is left unexplored.

An approach that bears some similarities to ours is the SpikeProp algorithm~\cite{Bohte_etal02} that can be used to train multi-layer spiking networks to produce output spikes at specific times. SpikeProp assumes a connection between two spiking neurons consists of a number of sub-connections, each with a different delay and a trainable weight. We use a  more conventional network model that does not depend on combinations of pre-specified delay elements to transform input spike times to output spike times, and instead relies only on simple neural and synaptic dynamics. Unlike SpikeProp, our formulation results in an analytical relation between input and output spike times. The spiking network thus does not have to be simulated during the training loop. This accelerates training and allows us to make use of standard GPU-accelerated ANN training packages to scale the training to larger datasets.    

Many approaches for training spiking networks first train conventional feedforward ANNs and then translate the trained weights to spiking networks developed to approximate the behavior of the original ANNs~\cite{OConnor_etal13,Diehl_etal15,Cao_etal15}. The spiking networks obtained using these methods use rate coding where the spiking rate of a neuron encodes an analog quantity corresponding to the analog output of an ANN neuron. High spike rates would then mask the discrete nature of the spiking activity.
 
In this paper, we develop a direct training approach that does not try to reduce spiking networks to conventional ANNs. Instead, we relate the time of any spike differentiably to the times of all spikes that had a causal influence on its generation. We can then impose any differentiable cost function on the spike times of the network and minimize this cost function directly through gradient descent. 

By using spike times as the information-carrying quantities, we avoid having to work with discrete spike counts or spike rates, and instead work with a continuous representation (spike times) that is amenable to gradient descent training. This training approach allows detailed control over the behavior of the network (at the level of single spike times) which would not be possible in training approaches based on rate-coding.

Since we use a temporal spike code, neuron firing can be quite sparse as the time of each spike carries significant information. Compared to rate-based networks, the networks we present can  be implemented more efficiently on neuromorphic architectures where power consumption decreases as spike rates are reduced~\cite{Qiao_etal15,Neil_Liu14,Benjamin_etal14}. %Neuromorphic processing architectures built using spiking neurons promise improved performance in power-critical application due to the use of an event-driven massively parallel processing scheme~\cite{Merolla_etal14c}.
In conventional ANNs, each neuron calculates a weighted sum of the activities of all its source neurons then produces its output by applying a static non-linearity to this sum. We show that the behavior of the networks we present deviates quite significantly from this conventional ANN paradigm as the output of each neuron in one layer depends on a different and dynamically changing subset of the neurons in the preceding layer. Unlike rate-based spiking networks, the proposed networks can be directly trained using gradient descent methods, as time is the principal coding dimension, not the discrete spike counts. Unlike previous temporal learning approaches, our method extends naturally to multi-layer networks, and depends on deterministic neural and synaptic dynamics to optimize the spike times, rather than explicit delay elements in the network.

%% , but which can still be directly and effectively trained using standard gradient descent methods. The gradient descent methods directly optimize the time of each spike across multiple layer to minimize the training cost function which enables training of multi-layer networks. Gradient descent is applied directly,   m Unlike rate-based networks,  They resulting  the   Due to the use of neuronal temporal dynamics for classification, 

\section{Network model}
We use non-leaky integrate and fire neurons with exponentially decaying synaptic current kernels. The neuron's membrane dynamics are described by:
\begin{equation}
\label{eq:model_neuron}
\frac{dV^j_{mem}(t)}{dt} = \sum\limits_i w_{ji}\sum\limits_r\kappa(t-t_i^r)
\end{equation}
where $V^j_{mem}$ is the membrane potential of neuron $j$. The right hand side of the equation is the synaptic current. $w_{ji}$ is the weight of the synaptic connection from neuron $i$ to neuron $j$ and $t_i^r$ is the time of the $r^{th}$ spike from neuron $i$. $\kappa$ is the synaptic current kernel given by:
\begin{equation}
\label{eq:model_synapse}
\kappa(x) = \Theta(x)exp(-\frac{x}{\tau_{syn}}) \quad \text{where} \quad \Theta(x) = \begin{cases}
   1 & \text{if \quad $x\geq 0$}  \\
   0       & \text{otherwise} 
  \end{cases}
\end{equation}
Synaptic current thus jumps instantaneously on the arrival of an input spike, then decays exponentially with time constant $\tau_{syn}$. Since $\tau_{syn}$ is the only time constant in the model, we set it to 1 in the rest of the paper, i.e, normalize all times with respect to it. The neuron spikes when its membrane potential crosses a firing threshold which we set to $1$, i.e, all synaptic weights are normalized with respect to the firing threshold. The membrane potential is reset to $0$ after a spike. We allow the membrane potential to go below zero if the integral of the synaptic current is negative.

Assume a neuron receives $N$ spikes at times $\{t_1,..,t_N\}$ with weights $\{w_1,..,w_N\}$ from $N$ source neurons. Each weight can be positive or negative. Assume the neuron spikes in response at time $t_{out}$. By integrating Eq.~\ref{eq:model_neuron}, the membrane potential for $t < t_{out}$ is given by:
\begin{equation}
\label{eq:tfire1}
V_{mem}(t) = \sum\limits_{i=1}^N \Theta(t-t_i) w_i (1-exp(-(t-t_i)))
\end{equation}
Assume only a subset of these input spikes with indices in $C \subseteq \{1,..,N\}$ had arrived before $t_{out}$ where $C= \{i : t_i < t_{out}\}$. It is only these input spikes that influence the time of the output neuron's first spike. We call this set of input spikes the causal set of input spikes. The sum of the weights of the causal input spikes has to be larger than 1, otherwise they could not have caused the neuron to fire. From Eq.~\ref{eq:tfire1}, $t_{out}$ is then implicitly defined as:
\begin{equation}
1 = \sum\limits_{i \in C} w_i (1-exp(-(t_{out}-t_i)))
\end{equation} 
where $1$ is the firing threshold. Hence,
\begin{equation}
exp(t_{out}) = \frac{\sum\limits_{i \in C} w_i exp(t_i)}{\sum\limits_{i \in C} w_i -1}
\end{equation} 
Spike times always appear exponentiated. Therefore, we do a transformation of variables ${exp(t_x) \rightarrow z_x}$  yielding an expression relating input spike times to the time of the first spike of the output neuron in the post-transformation domain (which we denote as the z-domain):
\begin{equation}
\label{eq:master}
z_{out} = \frac{\sum\limits_{i \in C} w_i z_i}{\sum\limits_{i \in C} w_i -1}
\end{equation}
Note that for the neuron to spike in the first place, we must have $\sum\limits_{i \in C} w_i  > 1$, so $z_{out}$ is always positive (one can show this is the case even if some of the weights are negative). It is also always larger than any element of $\{z_{i} : i \in C\}$, i.e, the output spike time is always larger than any input spike time in the causal set which follows from the definition of the causal set. We can obtain a similar expression relating the time of the $Q^{th}$ spike of the output neuron, $z_{out}^Q$, to the input spike times in the z-domain:
\begin{equation}
\label{eq:subsequent_spikes}
z_{out}^Q = \frac{\sum\limits_{i \in C^Q} w_i z_i}{\sum\limits_{i \in C^L} w_i - Q}
\end{equation}
where $C^Q$ is the set of indices of the input spikes that arrive before the $Q^{th}$ output spike. Equation~\ref{eq:subsequent_spikes} is only valid if the denominator is positive, i.e, there are sufficient input spikes with large enough total positive weight to push the neuron past the firing threshold $Q$ times. In the rest of the paper, we consider a neuron's output value to be the time of its first spike. Moreover, once a neuron spikes, it is not allowed to spike again, i.e, we assume it enters an infinitely long refractory period. We allow each neuron to spike at most once for each input presentation in order to make the spiking activity sparse and to force the training algorithm to make optimal use of each spike. During training, we use a weight cost term that insures the neuron receives sufficient input to spike as we describe in the next section.

\begin{figure*}[t]
\centering
  \centering
  \begin{subfigure}[b]{0.35\textwidth}
    \includegraphics[width=\textwidth]{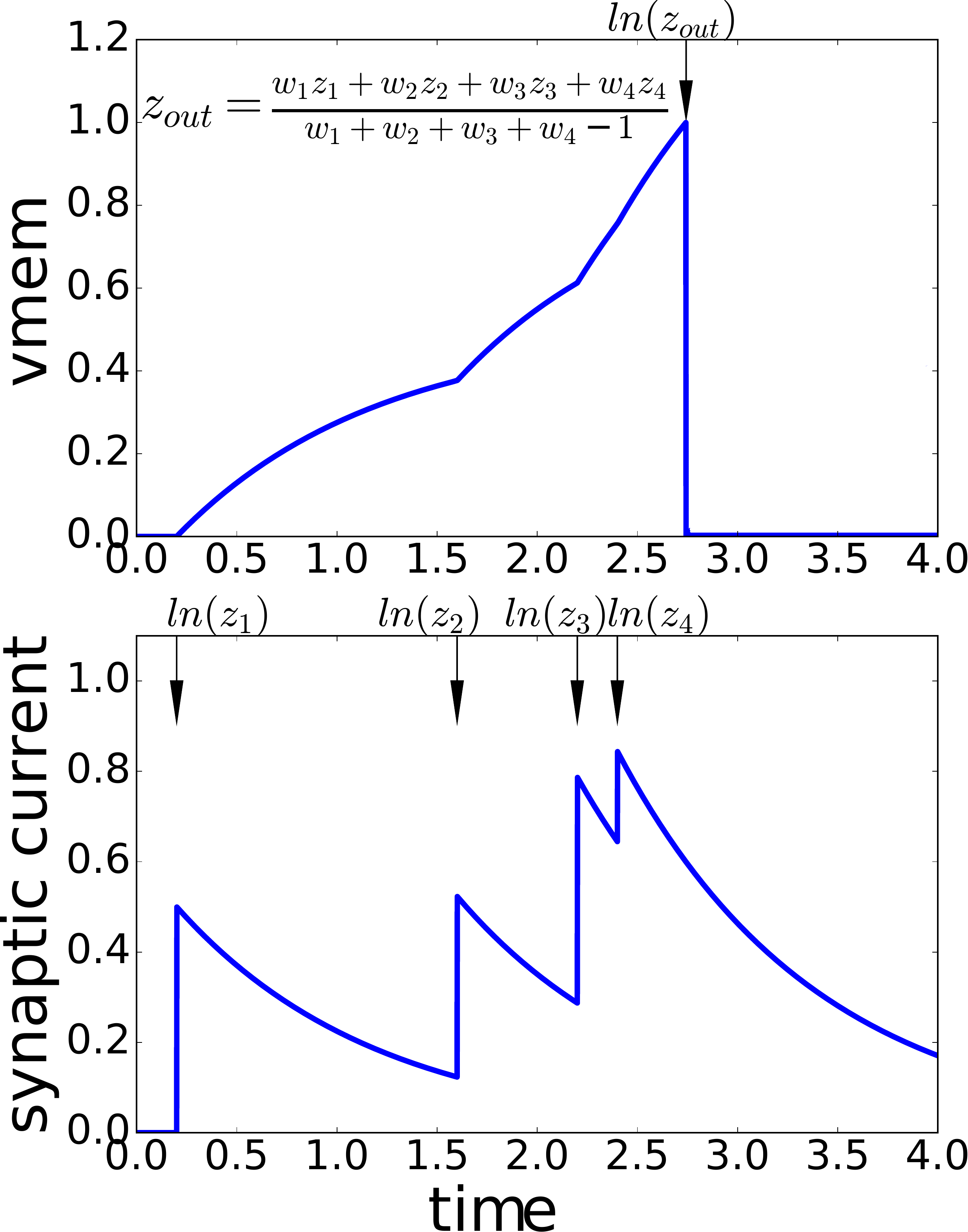}
    \subcaption{}
    \label{fig:causal_a}
  \end{subfigure}
  \quad
  \begin{subfigure}[b]{0.35\textwidth}
    \includegraphics[width=\textwidth]{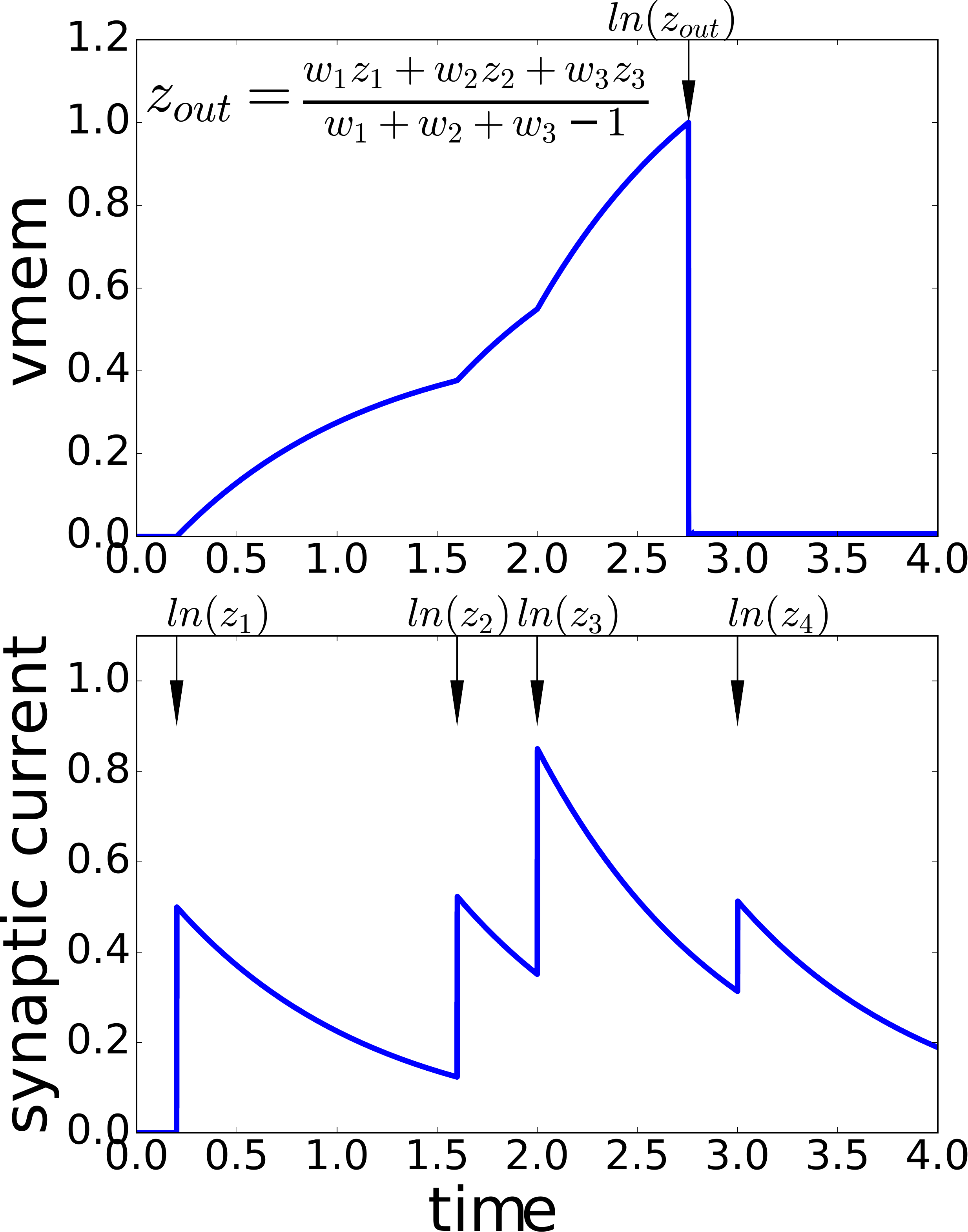}
    \subcaption{}
    \label{fig:causal_b}
  \end{subfigure}

\caption{Changes in the causal input set modify the linear input-output spike times relation. Plots show the synaptic current and the membrane potential of a neuron in two situations: in (\subref{fig:causal_a}), the neuron receives 4 spikes with weights $\{w_1,w_2,w_3,w_4\}$ at the times indicated by the black arrows in the bottom plot. This causes the neuron to spike at time $t_{out} = ln(z_{out})$; in (\subref{fig:causal_b}), input spike times change causing the neuron to spike before the fourth input spike arrives. This changes the linear z-domain input-output relation of the neuron compared to (\subref{fig:causal_a}). Note that a neuron is only allowed to spike once, after which it can not spike again until the network is reset and a new input pattern is presented.} 
\label{fig:causal}
\end{figure*}

The linear relation between input and output spike times in the z-domain is only valid in a local interval. A different linear relation holds when the set of causal input spikes changes. This is illustrated in Fig.~\ref{fig:causal}, where the 4th input spike is part of the causal set in one case but not in the other. From Eq.~\ref{eq:master}, the effective weight of input $z_p$ in the linear input-output relation in the z-domain is $w_p/(\sum\limits_{i \in C} w_i -1)$. This effective weight depends on the weights of the spikes in the causal set of input spikes. As this causal set changes due to the changing spike times from the source neurons, the effective weight of the different input spikes that remain in the causal set changes (as is the case for the first three spikes in Figs. ~\ref{fig:causal_a} and \ref{fig:causal_b} whose effective weight changes as the causal set changes, even though their actual synaptic weights are the same ). For some input patterns, some source neurons may spike late causing their spikes to leave the causal set of the output neuron and their effective spike weights to become zero. Other source neurons may spike early and influence the timing of the output neuron's spike and thus their spikes acquire a non-zero effective weight.

The causal set of input spikes is dynamically determined based on the input spike times and their weights. Many early spikes with strong positive weights will cause the output neuron to spike early, negating the effect of later spikes on the output neuron's first spike time regardless of the weights of these later spikes. The non-linear transformation from ${\bf z} = \{z_1,..,z_N\}$ to $z_{out}$ implemented by the spiking neuron is thus fundamentally different from the static non-linearities used in traditional ANNs where only the aggregate weighted input is considered. 

The non-linear transformation implemented by the spiking neuron is continuous in most case, i.e, small perturbation in  ${\bf z}$ will lead to proportionately small perturbations in $z_{out}$. This is clear when the perturbations do not change the set of causal input spikes as the same linear relation continues to hold. Consider, however, the case of an input spike with weight $w_x$ that occurs just after the output spike at time $z_x = z_{out}+\epsilon$. A small perturbation pushes this input spike to time $z_x = z_{out}-\epsilon$ adding it to the causal set. By applying Eq.~\ref{eq:master}, the perturbed output time is ${z_{out}^{perturb} = (\sum\limits_{i \in C} w_i z_i + w_x z_x)/(\sum\limits_{i \in C} w_i + w_x -1)}$ where $C$ is the causal set before the perturbation. Substituting for $z_x$, ${z_{out}^{perturb} - z_{out} = -\epsilon w_x/(\sum\limits_{i \in C} w_i + w_x -1)}$. The output perturbation is thus proportional to the input perturbation but this is only the case when $\sum\limits_{i \in C} w_i + w_x > 1$, otherwise the perturbed input spike with negative weight at $z_{out}-\epsilon$ would cancel the original output spike at $z_{out}$. In summary, the input spike times to output spike time transformation of the spiking neuron is continuous except in situations where small perturbations affect whether a neuron spikes or not.

The fact that we get a piece-wise linear relation between input and output spike times in the z-domain is a general phenomenon and not an accident of our model. Scaling the input vector ${\bf z} = [z_1,..,z_N]$ by a factor $K$ is equivalent to shifting all input spike times in the time domain, $\{t_1,..,t_N\}$, forward by time $ln(K)$. This would shift the output spike time, $t_{out}$, forward by time $ln(K)$ as well since the network has no internal time reference and shifting all input spikes in time would thus shift all output spikes by the same amount. This would in turn linearly scale $z_{out}$ by a factor $K$, realizing a (locally) linear input-output relation.

\section{Training}
\label{sec:training}
We consider feedforward neural networks where the neural and synaptic dynamics are described by Eqs.~\ref{eq:model_neuron} and ~\ref{eq:model_synapse}. The neurons are arranged in a layer-wise manner. Neurons in one layer project in an all-to-all fashion to neurons in the subsequent layer. A neuron's output is the time of its first spike and we work exclusively in the z-domain. Once a neuron spikes, it is not allowed to spike again until the network is reset and a new input pattern is presented. The forward pass is described in Algorithm~\ref{alg:fforward}. $z^{r}$ is the vector of the first spike times from each neuron in layer $r$.  All indices are 1-based, vectors are in boldface, and the $i^{th}$ entry of a vector ${\bf a}$ is ${\bf a}[i]$. At each layer, the causal set for each neuron is obtained from the $get\_causal\_set$ function. The neuron output is then evaluated according to Eq.~\ref{eq:master}. The $get\_causal\_set$ function is shown in algorithm~\ref{alg:get_causal}. It first sorts the input spike times, and then considers increasingly larger sets of the early input spikes until it finds a set of input spikes that causes the neuron to spike. The output neuron must spike at a time that is less than the time of any input spike not in the causal set, otherwise the causal set is incomplete. If no such set exists, i.e, the output neuron does not spike in response to the input spikes, $get\_causal\_set$ returns the empty set $\Phi$ and the neuron's output spike time is set to infinity (maximum positive value in implementation).
\begin{algorithm}
\caption{Forward pass in a feedforward spiking network with L layers}
\label{alg:fforward}
\begin{algorithmic}[1]
\State \textbf{Input:} ${\bf z^0}$: Vector of input spike times
\State \textbf{Input:} $\{N^1,..,N^L\}$: Number of neurons in the L layers
\State \textbf{Input:} $\{W^1,..,W^L\}$: Set of weight matrices. $W^l[i,j]$ is the weight from neuron j in layer $l-1$ to neuron $i$ in layer $l$
\State \textbf{Output:} ${\bf z^L}$: Vector of first spike times of neurons in the top layer
\For {r= 1 to L}
\For {i= 1 to $N^r$}
\State $C_i^r \gets get\_causal\_set({\bf z^{r-1}},W^r[i,:])$
\If {$C_i^r \neq \Phi$}
\State ${\bf z^{r}}[i] \gets \frac{\sum\limits_{k \in C_i^r} W^{r}[i,k]{\bf z^{r-1}}[k]}{\sum\limits_{k \in C_i^r} W^{r}[i,k] -1}$
\Else
\State ${\bf z^{r}}[i] \gets \infty$
\EndIf
\EndFor
\EndFor 
%\State \textbf{return} ${\bf z^L}$
\end{algorithmic}
\end{algorithm}
\begin{algorithm}

\caption{get\_causal\_set: Gets indices of input spikes influencing first spike time of output neuron}
\label{alg:get_causal}
\begin{algorithmic}[1]
\State \textbf{Input:} ${\bf z}$: Vector of input spike times of length N
\State \textbf{Input:} ${\bf w}$: Weight vector of the input spikes
\State \textbf{Output:} $C$: Causal index set
\State ${\bf sort\_indices} \gets argsort({\bf z})$ //Ascending order argsort
\State ${\bf z^{sorted}} \gets {\bf z}[{\bf sort\_indices}]$ //sorted input vector
\State ${\bf w^{sorted}} \gets {\bf w}[{\bf sort\_indices}]$ //weight vector rearranged to match sorted input vector
\For {i= 1 to $N$}
\If {$i == N$}
\State $next\_input\_spike \gets \infty$
\Else
\State $next\_input\_spike \gets {\bf z^{sorted}}[i+1]$
\EndIf
\If {$\sum\limits_{k = 1}^i {\bf w^{sorted}}[k] > 1 \quad  \wedge \quad \frac{\sum\limits_{k=1}^i {\bf w^{sorted}}[k]{\bf z^{sorted}}[k]}{\sum\limits_{k = 1}^i {\bf w^{sorted}}[k] -1} < next\_input\_spike$}
\State \textbf{return} $\{{\bf sort\_indices}[1],..,{\bf sort\_indices}[i]\}$
\EndIf
\EndFor
\State \textbf{return} $\Phi$
\end{algorithmic}
\end{algorithm}

From Eq.~\ref{eq:master}, the derivatives of a neuron's first spike time with respect to synaptic weights and input spike times are given by: 

\begin{equation}
\label{eq:deriv1}
\frac{dz_{out}}{dw_p} = \begin{cases}
   \frac{z_p - z_{out}}{\sum\limits_{i \in C} w_i -1} & \text{if \quad $p\in C$}  \\
   0       & \text{otherwise} 
  \end{cases}
\end{equation}
\begin{equation}
\label{eq:deriv2}
\frac{dz_{out}}{dz_p} = \begin{cases}
   \frac{w_p}{\sum\limits_{i \in C} w_i -1} & \text{if \quad $p\in C$}  \\
   0       & \text{otherwise} 
  \end{cases}
\end{equation}

Unlike the spiking neuron's input-output relation which can still be continuous at points where the causal set changes, the derivative of the neuron's output with respect to inputs and weights given by Eqs.~\ref{eq:deriv1} and~\ref{eq:deriv2} is discontinuous at such points. 
This is not a severe problem for gradient descent methods. Indeed, many feedforward ANNs use activation functions with a discontinuous first derivative such as rectified linear units (ReLUs)~\cite{Nair_Hinton10} while still being effectively trainable. 

A differentiable cost function can be imposed on the spike times generated anywhere in the network. The gradient of the cost function with respect to the weights in lower layers can be evaluated by backpropagating errors through the layers using Eqs.~\ref{eq:deriv1} and~\ref{eq:deriv2} through the standard backpropagation technique. In the next section, we use the spiking network in a classification setting. In training the network, we had to use the following techniques to enable the networks to learn:
\paragraph{Constraints on synaptic weights} We add a term to the cost function that heavily penalizes neurons' input weight vectors whose sum is less than 1. During training, this term pushes the sum of the weights in each neuron's input weight vector above 1 which ensures that a neuron spikes if all its input neurons spike. This term in the cost function is crucial, otherwise the network can become quiescent and stop spiking. This cost term has the form
\begin{equation}
  \label{eq:weightsum}
  WeightSumCost = K*\sum_jmax(0,1 - \sum_i w_{ji})
\end {equation}
where the summation over $j$ runs over all neurons and the summation over $i$ runs over all the neurons that project to neuron $j$. $w_{ji}$ is the connection weight from neuron $i$ to neuron $j$. $K$ is a hyper-parameter. $K$ is typically chosen to be larger than $1$ to strictly enforce the constraint that the incoming weight vector to each neuron sums to more than one. Large positive weights are problematic as they can enable a source neuron to almost unilaterally control the target neuron's spike time, compromising the ability of the target neuron to integrate information from all its input neurons. Therefore, we use L2 weight regularization to stop weights from becoming too large. 

\paragraph{Gradient normalization} We observed that the gradients can become very large during training. This is due to the highly non-linear relation between the output spike time and the weights when the sum of the weights for the causal set of input spikes is close to 1. This can be seen from Eqs.~\ref{eq:master},~\ref{eq:deriv1}, and~\ref{eq:deriv2} where a small denominator can cause the output spike time and the derivatives to diverge. This hurts learning as it causes weights to make very large jumps. We use gradient normalization to counter that: if the Frobenius norm of the gradient of a weight matrix is above a threshold, we scale the gradient matrix so that its Frobenius norm is equal to the threshold before doing the gradient descent step. To reduce the dependence of this gradient normalization scheme on weight matrix size, we first normalize the Frobenius norm of the weight gradient matrix by the number of source neurons (number of rows in the weight matrix).

\section{Results}
We trained the network to solve two classification tasks: an XOR task, and the permutation invariant MNIST task~\cite{Le-Cun_etal98}. We chose the XOR task to illustrate how the network is able to implement a non-linear transformation since a linear network can not solve the XOR task. The MNIST task was chosen to examine the generalization behavior of the network as the network is tested on input patterns that it has never seen before. We used fully connected feedforward networks where the top layer had as many neurons as the number of classes (2 in the XOR task and 10 in the MNIST task). The goal is to train the network so that the neuron corresponding to the correct class fires first among the top layer neurons. We used the cross-entropy loss and interpreted the value of a top layer neuron as the negative of its spike time (in the z-domain). Thus, by maximizing the value of the correct class neuron value, training effectively pushes this neuron to fire earlier than the neurons representing the incorrect classes. For an output spike times vector ${\bf z}^L$ and a target class index $g$, the loss function is given by
\begin{equation}
L(g,{\bf z^L}) = -ln \frac{exp(-{\bf z^L}[g])}{\sum\limits_i exp(-{\bf z^L}[i])} 
\end{equation} 

We used standard gradient descent to minimize the loss function across the training examples. 
Training was done using Theano~\cite{Bastien_etal12,Bergstra_etal10}.
\subsection{XOR task}
\begin{figure*}[t]
\centering
  \centering
  \begin{subfigure}[b]{0.3\textwidth}
    \includegraphics[width=\textwidth]{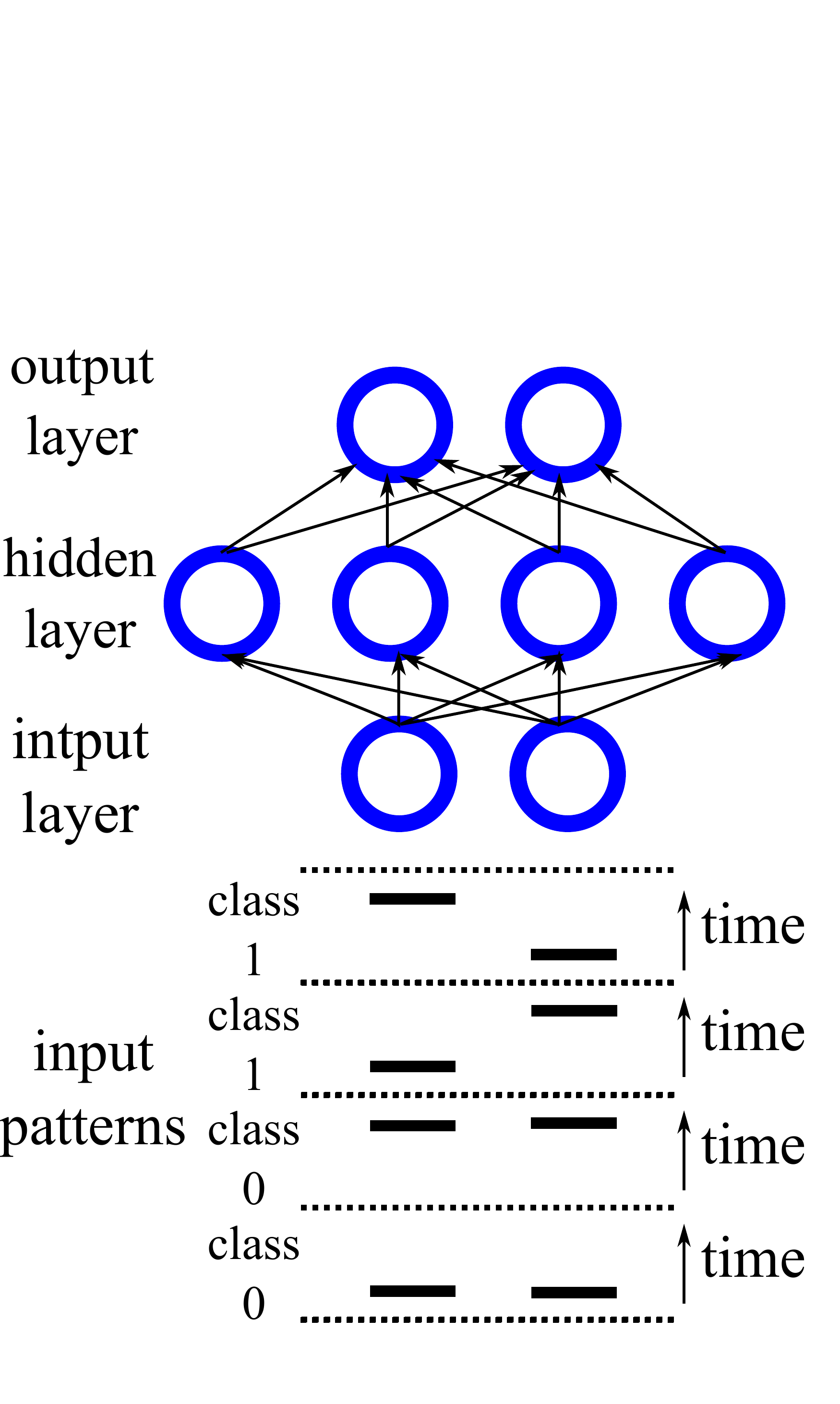}
    \subcaption{}
    \label{fig:xor_a}
  \end{subfigure}
  \quad
  \begin{subfigure}[b]{0.65\textwidth}
    \includegraphics[width=\textwidth]{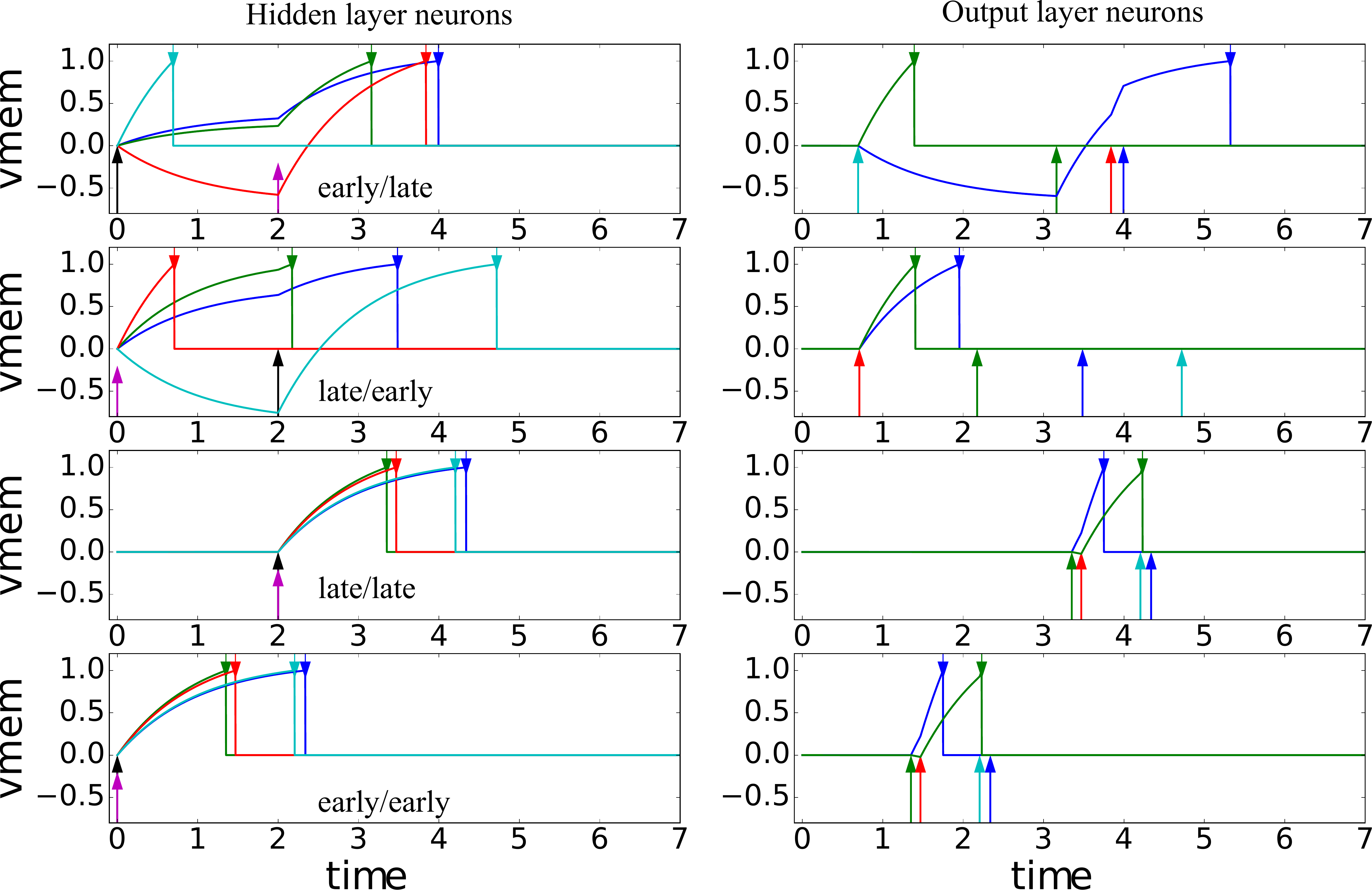}
    \subcaption{}
    \label{fig:xor_b}
  \end{subfigure}
\caption{(\subref{fig:xor_a}) Network implementing the XOR task using one hidden layer. There are 4 input patterns divided into two classes. (\subref{fig:xor_b}) Post-training simulation results of the network in (\subref{fig:xor_a}) for the four input patterns, one pattern per row. The left plots show the membrane potential of the four hidden layer neurons, while the right plots show the membrane potential of the two output layer neurons. Arrows at the bottom of the plot indicate input spikes to the layer, while arrows at the top indicate output spikes. The output spikes of the hidden layer are the input spikes of the output layer. The input pattern is indicated by the text in the left plots, and also by the pattern of input spikes.} 
\end{figure*}

In the XOR task, two spike sources send a spike each to the network. Each of the two input spikes can occur at time 0.0 (early spike) or 2.0 (late spike). The two input spike sources project to a hidden layer of 4 neurons and the hidden neurons project to two output neurons. The first output neuron must spike before the second output neuron if exactly one input spike is an early spike. The network is shown in Fig.~\ref{fig:xor_a} together with the 4 input patterns.

To investigate whether the network can robustly learn to solve the XOR task, we repeated the training procedure 1000 times starting from random initial weights each time. In each of these 1000 training trials, we used as many training iterations as needed for training to converge. We used a constant learning rate of $0.1$. The weight sum cost coefficient ($K$ in Eq.~\ref{eq:weightsum}) is $10$. We did not use L2 regularization. The maximum allowed row-normalized Frobenius norm of the gradient of a weight matrix is $10$.   % and the network to classify all input patterns correctly. 
 Each training iteration involved presenting the four input patterns 100 times. Across the 1000 trials, the maximum number of training iterations needed to converge was $61$ while the average was $3.48$. 
%i.e, on average, the network needed 348 presentations of the 4 input patterns to converge. 
Figure~\ref{fig:xor_b} shows the post-training simulation results of the network when presented with each of the input patterns. The causal input sets of the different neurons change across the input patterns, allowing the network to implement the non-linearity needed to solve the XOR task.

\subsection{MNIST classification task}
The MNIST database contains 70,000 28x28 grayscale images of handwritten digits. The training set of 60,000 labeled digits was used for training, and testing was done using the remaining 10,000. No validation set was used. All grayscale images were first binarized to two intensity values: high and low. Pixels with high intensity generate a spike at time $0$, while pixels with low intensity generate a spike at time $ln(6) = 1.79$, corresponding to $z=6$ in the z-domain. $ln(6) = 1.79$ was chosen to provide a large enough temporal separation between spikes from high intensity pixels and spikes from low intensity pixels. We noticed that accuracy suffered if this temporal separation was decreased below the synaptic time constant, i.e, below a time unit of $1$, while increasing temporal separation further did not have an appreciable effect on accuracy. All times are normalized with respect to the synaptic time constant (see Eq.~\ref{eq:model_synapse}). We investigated two feedforward network topologies with fully connected hidden layers: the first network has one hidden layer of 800 neurons (the 784-800-10 network), and the second has two hidden layers of 400 neurons each (the 784-400-400-10 network). We found that accuracy is slightly improved if we use an extra reference neuron that always spikes at time 0 and projects through trainable weights to all neurons in the network. We ran 100 epochs of training with an exponentially decaying learning rate. We tried different learning rates and fastest convergence was obtained when learning rate starts at $0.01$ in epoch 1 and ends at $0.0001$ in epoch 100. We used a mini-batch size of 10. The weight sum cost coefficient ($K$ in Eq.~\ref{eq:weightsum}) is $100$. The L2 regularization coefficient is $0.001$. The maximum allowed row-normalized Frobenius norm of the gradient of a weight matrix is $10$. Each of the two network topologies was trained twice, once with non-noisy input spike times and once with noise-corrupted input spike times. In the noisy input case, noise delays each spike with the absolute value of a random quantity drawn from a zero mean, unity variance Gaussian distribution. Noise was only used during training.
Table~\ref{table:mnist} shows the performance results for the two networks after the noisy and non-noisy training regimes. 

The small errors on the training set indicate the networks have enough representational power to solve this task, as well as being effectively trainable. The networks overfit the training set as indicated by the significantly higher test set errors. Noisy training input helps in regularizing the networks as it reduces test set error but further regularization is still needed. We experimented with dropout~\cite{Srivastava_etal14} where we randomly removed neurons from the network during training. However, dropout does not seem to be a suitable technique in our networks as it reduces the number of spikes received by the neurons, which would often prevent them from spiking. Effective techniques are still needed to combat overfitting and allow better generalization in the proposed networks.

\begin{table}[t]
  \caption{Performance results for the permutation-invariant MNIST task}
  \label{table:mnist}
  \centering
  \begin{tabular}[t]{lll}
    \toprule
    Network     & Training error  & Test error    \\
    \midrule
    784-800-10 (non-noisy training input)  & 0.013\%   & 2.8\%     \\
    784-800-10 (noisy training input)     & 0.005\% & 2.45\%      \\
    784-400-400-10 (non-noisy training input) & 0.031\%  & 3.08\%      \\
    784-400-400-10 (noisy training input)     & 0.255\% & 2.86\%      \\

    \bottomrule
  \end{tabular}
\end{table}

\begin{figure*}[t]
\centering
  \centering
  \begin{subfigure}[b]{0.39\textwidth}
    \includegraphics[width=\textwidth]{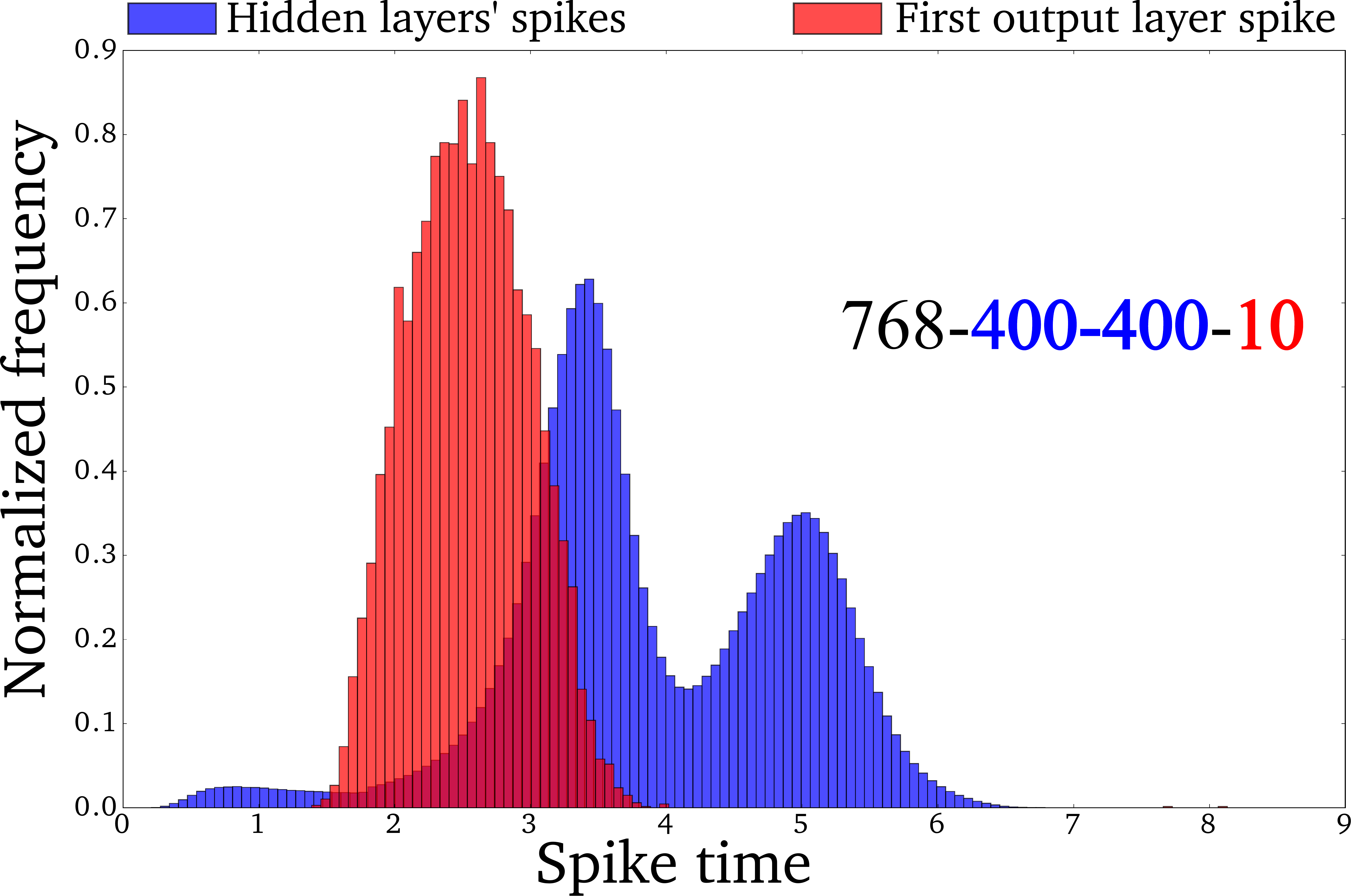}
    \subcaption{}
    \label{fig:mnist_a}
  \end{subfigure}
  \quad
  \begin{subfigure}[b]{0.39\textwidth}
    \includegraphics[width=\textwidth]{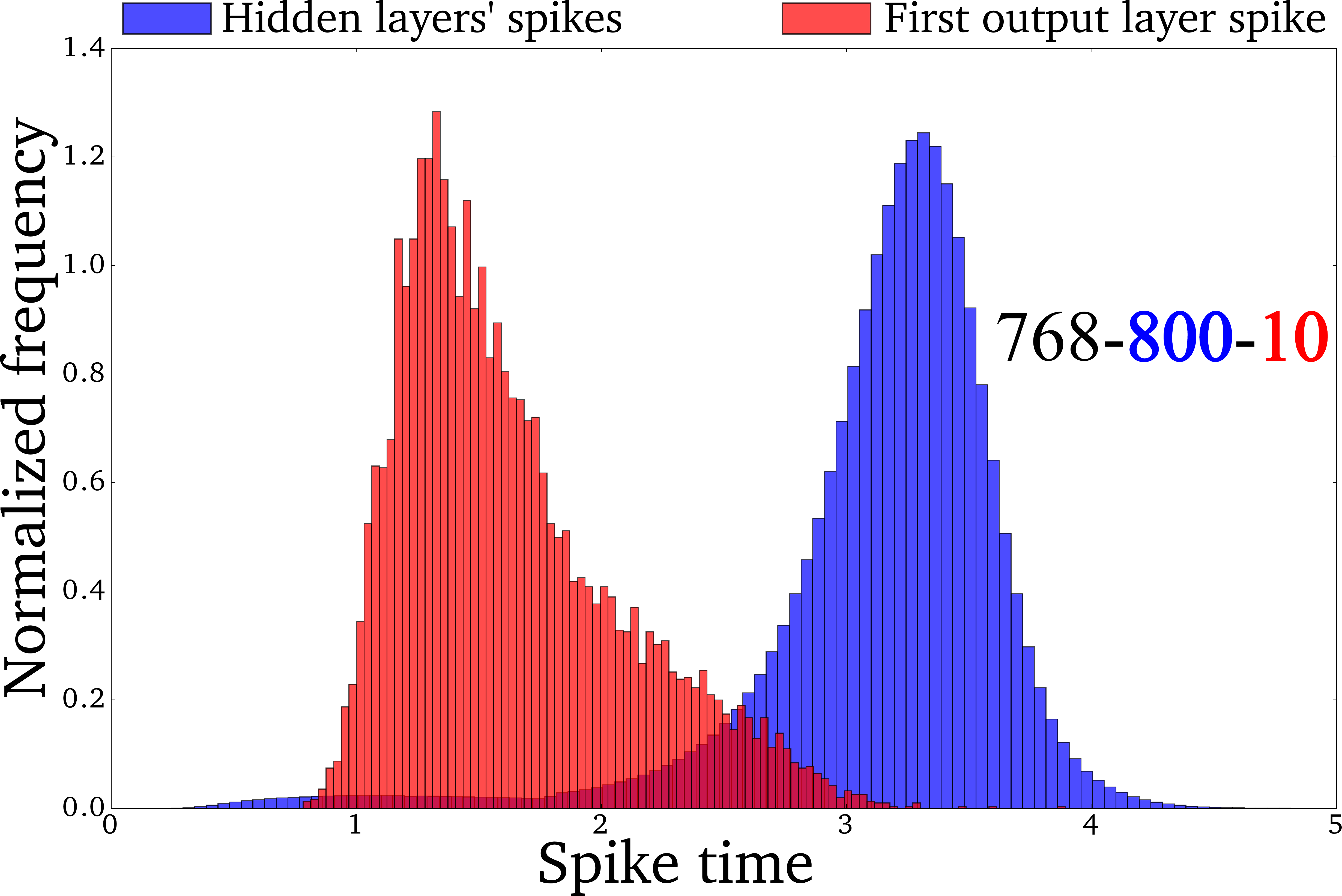}
    \subcaption{}
    \label{fig:mnist_b}
  \end{subfigure}\\
  \begin{subfigure}[b]{0.39\textwidth}
    \includegraphics[width=\textwidth]{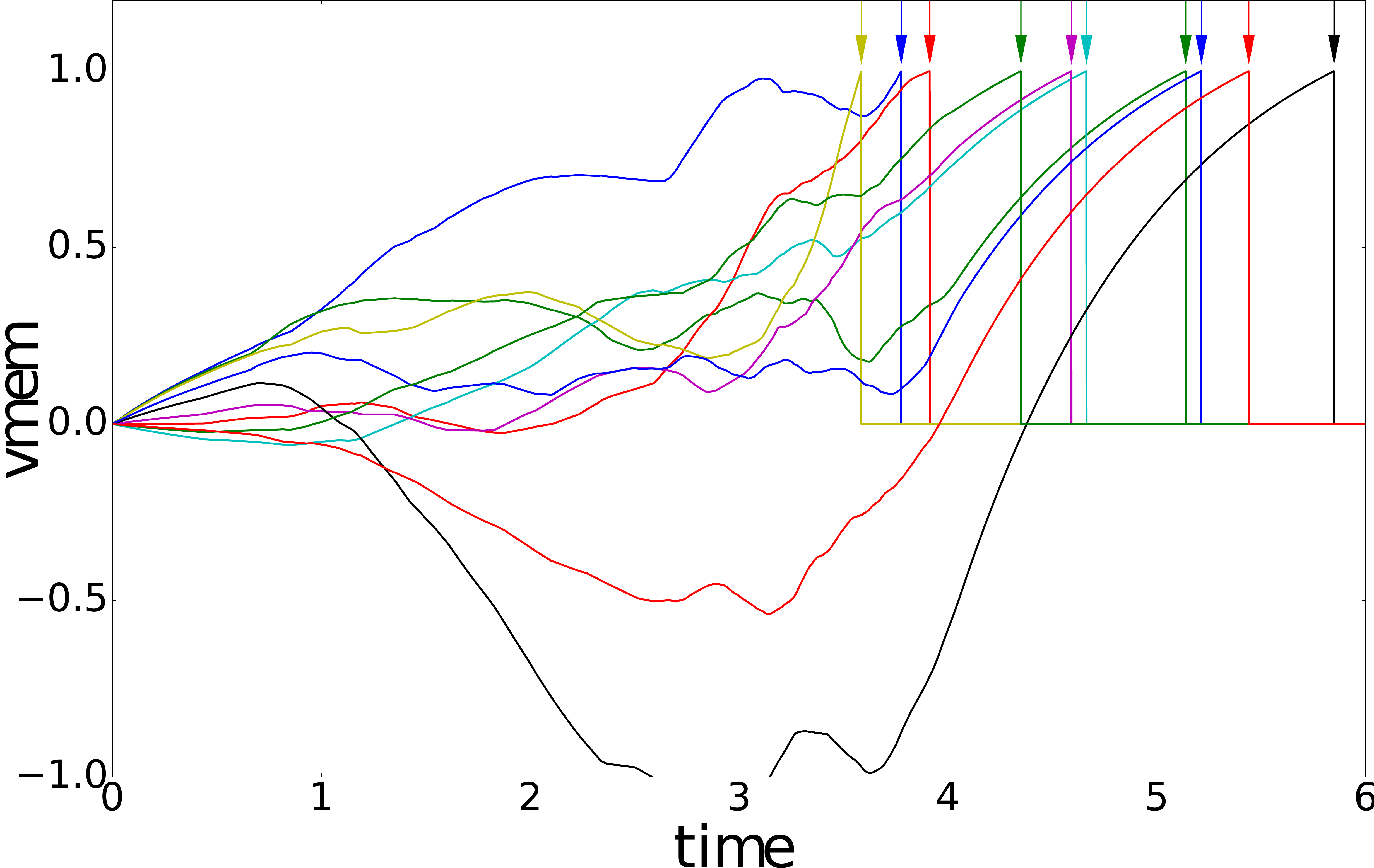}
    \subcaption{}
    \label{fig:mnist_c}
  \end{subfigure}
  \quad
  \begin{subfigure}[b]{0.39\textwidth}
    \includegraphics[width=\textwidth]{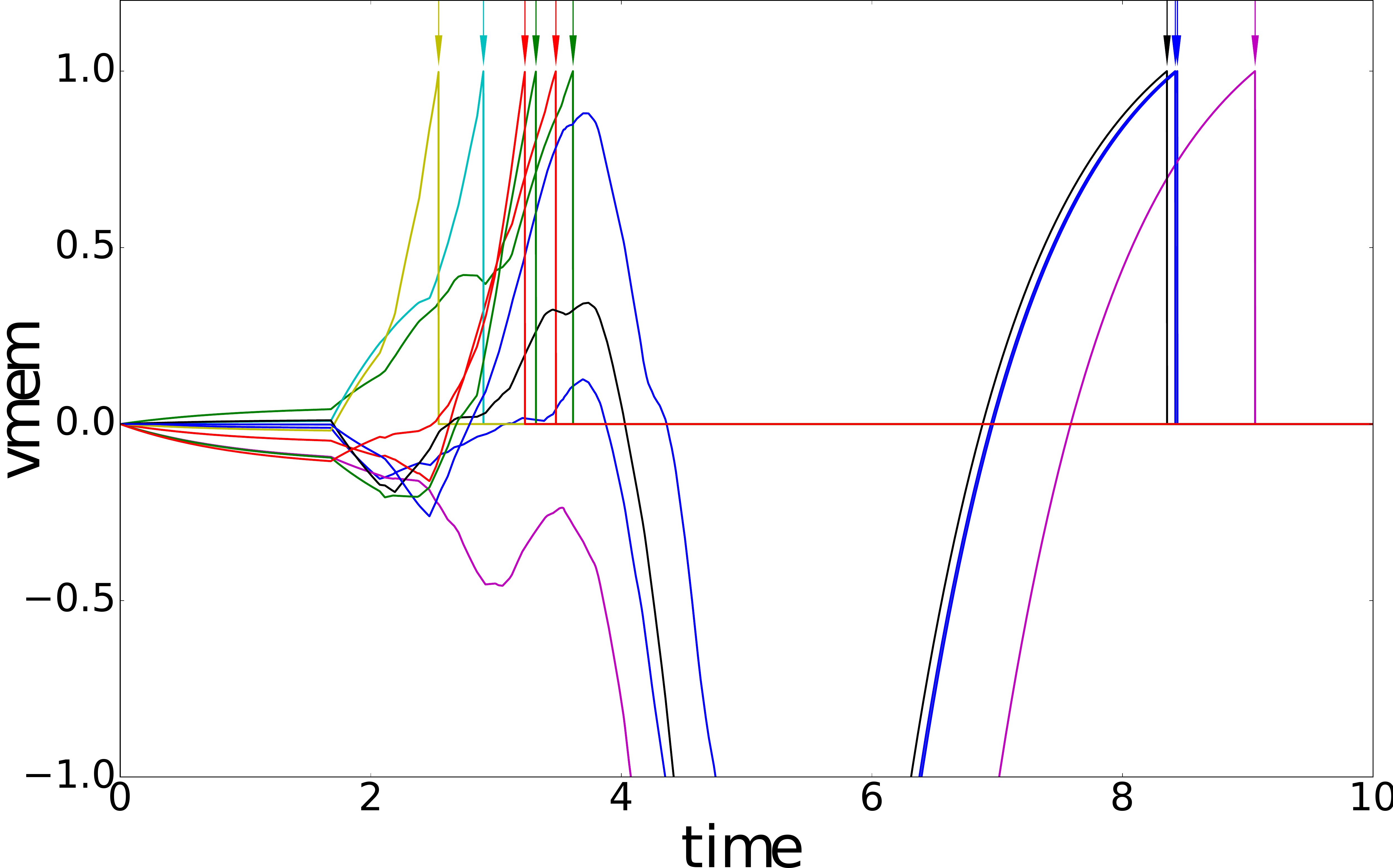}
    \subcaption{}
    \label{fig:mnist_d}
  \end{subfigure}

\caption{(\subref{fig:mnist_a},\subref{fig:mnist_b}) Histograms of spike times in the hidden layers and of the time of the first output layer spike across the 10,000 test set images for two network topologies. Both networks generate an output spike (i.e, select a class) before most of the hidden layer neurons have managed to spike. (\subref{fig:mnist_c}) Evolution of the membrane potentials of 10 neurons in the second hidden layer of the 768-400-400-10 network in response to a sample image. Top arrows indicate the spike (threshold crossing) times. (\subref{fig:mnist_d}) Evolution of the membrane potentials of the 10 output neurons in the same network and for the same input image as in (\subref{fig:mnist_c}). The earliest output neuron spikes (network selects a class) at time 2.5, i.e, before any of the 10 hidden neurons shown in (\subref{fig:mnist_c}) have spiked.} 
\label{fig:mnist}
\end{figure*}

\begin{figure*}[t]
\centering
\includegraphics[width=0.6\textwidth]{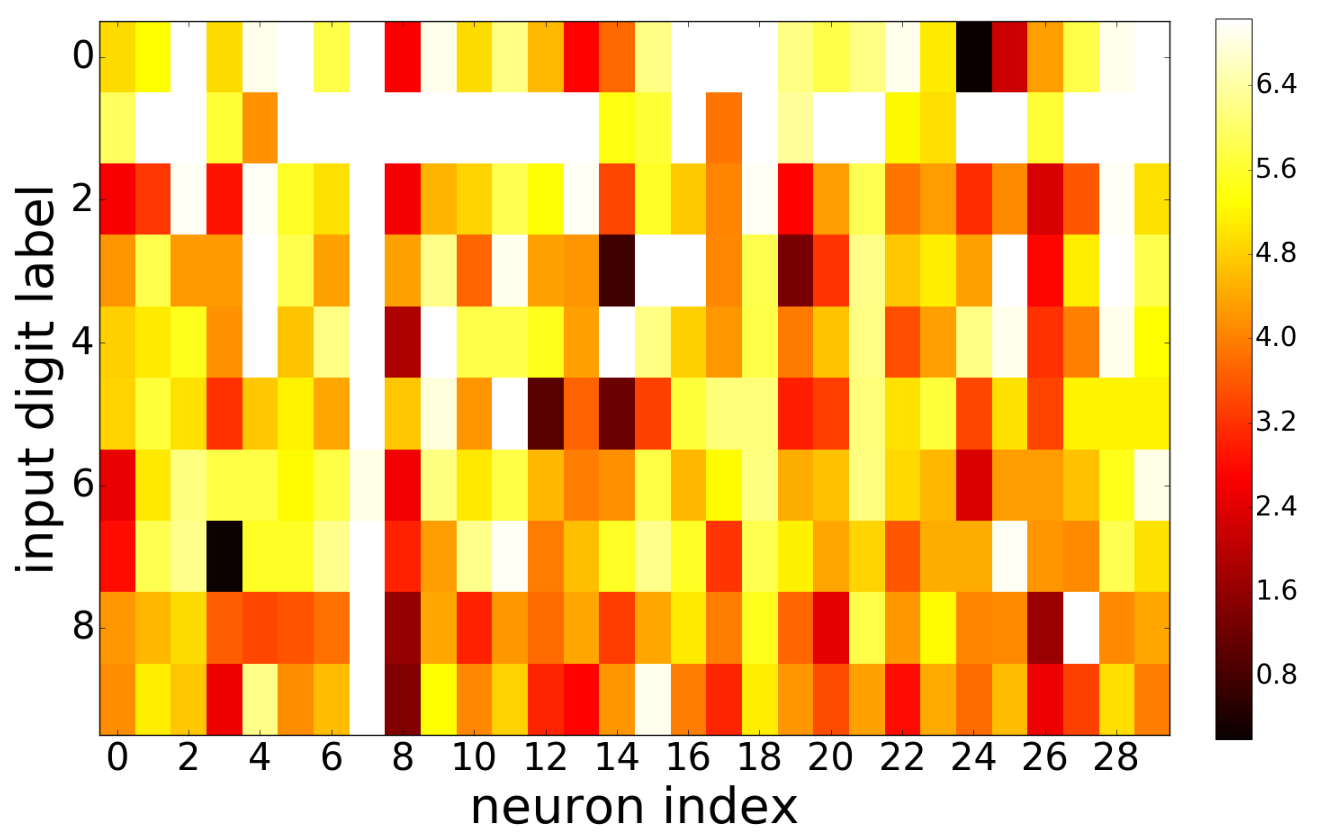}
\caption{Selectivity of 30 randomly selected hidden neurons in the 784-800-10 network to the 10 MNIST input classes. The plot shows the negative log probability for each of the 30 neurons to spike before the output layer spikes for each of the 10 input classes. This is the negative log probability that a neuron participates in the classification of a particular class. Probability was obtained from the network's response to the 10,000 test digits. Some neurons are highly selective. For example, neuron 3 is highly selective to the '7' digits while most of the neurons are more broadly tuned. Some rare neurons are mostly silent such as neuron 7, yet all neurons contribute to the classification of at least one of the 10,000 test patterns.} 
\label{fig:selectivity}
\end{figure*}

Figures~\ref{fig:mnist_a} and~\ref{fig:mnist_b} show the distribution of spike times in the hidden layers and the distribution of the times of the earliest output layer spike in the two networks. The later are the times at which the networks made a decision for the 10,000 test examples. Both networks were first trained using noisy input. For both topologies, the network makes a decision after only a small fraction of the hidden layer neurons have spiked. For the 784-800-10 topology, an output neuron spikes (a class is selected) after only 3.0\% of the hidden neurons have spiked (on average across the 10,000 test set images), while for the 784-400-400-10 topology, this number is 9.4\%. The network is thus able to make very rapid decisions about the input class, after approximately 1-3 synaptic time constants from stimulus onset, based only on the spikes of a small subset of the hidden neurons. This is illustrated in Figs.~\ref{fig:mnist_c} and~\ref{fig:mnist_d} which show the membrane potentials of 10 hidden neurons and the 10 output neurons. The spikes of the 10 hidden neurons do not factor into the network decision in this case as they all spike after the earliest output spike, i.e, after the network has already selected a class. 

Figure~\ref{fig:selectivity} shows the tuning properties of 30 randomly selected hidden layer neurons in the 784-800-10 network. We consider a hidden neuron to be tuned to a particular input class if it contributes to the classification of that class, i.e, if it spikes before the output layer spikes for that class. As shown in Fig.~\ref{fig:selectivity}, neurons are typically broadly tuned and contribute to the classification of many classes. No hidden layer neuron is redundant, i.e, no neuron can be removed from the network without affecting the output spike times across the MNIST test set.

\section{Conclusion}
We presented a form of spiking neural networks that can be effectively trained using gradient descent techniques. By using a temporal spike code, many difficulties involved in training spiking networks such as the discontinuous spike generation mechanism and the discrete nature of spike counts are avoided. The network input-output relation is piece-wise linear after a transformation of the time variable. As the input spike times change, the causal input sets of the neurons change, which in turn changes the form of the linear input-output relation (Fig.~\ref{fig:causal}). This is analogous to the behavior of networks using rectified linear units (ReLUs)~\cite{Nair_Hinton10} where changes in the input change the set of ReLUs producing non-zero output, thus changing the linear transformation implemented by the network. Piece-wise linear transformations can approximate any non-linear transformation to arbitrary accuracy~\cite{Goodfellow_etal13}. ANNs based on ReLU networks are currently setting the state of the art in various machine vision tasks~\cite{He_etal16,Huang_etal16}. As far as we know, this is the first time spiking networks have been shown to effectively implement a piece-wise linear transformation from input to output spike times.

We used standard stochastic gradient descent(SGD) during training. While second order methods~\cite{Martens10} could in principle be used, they are more computationally demanding than first-order methods like SGD. Furthermore, by augmenting the first order gradient information with various pieces of information about the gradient history~\cite{Kingma_Ba14}, the performance gap between first and second order methods can be eliminated in many cases~\cite{Sutskever_etal13}.

Recordings from higher visual areas in the brain indicate these areas encode information about abstract features of visual stimuli as early as $125\,ms$ after stimulus onset~\cite{Huang_etal05}. This is consistent with behavioral data showing response times in the order of a few hundred milliseconds in visual discrimination tasks~\cite{Fabre_etal98}. Given the typical firing rate of cortical neurons and delays across synaptic stages from the retina to higher visual areas, this indicates rapid visual processing is mostly a feedforward process where neurons get to spike at most once~\cite{Thorpe_etal01}.  The presented networks follow a similar processing scheme and could thus be used as a  trainable model to investigate the accuracy-response latency tradeoff in feedforward spiking networks. Output latency can be reduced by using a penalty term in the cost function that grows with the output spikes latency. Scaling this penalty term controls the tradeoff between minimizing latency and minimizing error during training. We used non-leaky integrate and fire neurons in our networks in order to obtain a closed form analytical expression relating input and output spike times. Biological neurons, on the other hand, have various leak mechanisms, allowing them to forget past subthreshold activity. A mechanism similar to leak-induced forgeting also occurs in our networks: information about the timing of incoming spikes is lost when the synaptic current for these spikes has decayed; the membrane potential indeed changes to a new value, but this value is independent of the ordering of past input spikes. Effective discrimination between different input temporal patterns can thus only happen when input spikes are within a few synaptic time constants of each other, which is also the case for leaky neurons.  

The performance of our network on the MNIST task falls short of the state of the art. Feedforward fully-connected ANNs achieve error rates between $0.9\%$ and $2\%$~\cite{Srivastava_etal14} on the MNIST task. ANNs, however, are evaluated layer by layer and can not be trained to produce a classification decision as soon as possible like the networks we describe in this paper. The training pressure to produce the classification result as soon as possible forces the described networks to ignore the majority of hidden neurons' activity by producing an output spike before most of the hidden neurons have spiked, which could explain the reduced accuracy. Rate-based spiking networks with a similar architecture to ours could achieve error rates as low as $1.3\%$~\cite{Lee_etal16}. They, however, make use of thousands of spikes that are integrated over time in order to yield an accurate classification result.

We considered the case where each neuron in the network is allowed to spike once. The training scheme can be extended to the case where each neuron spikes multiple times. The time of later spikes can be differentiably related to the times of all causal input spikes (see Eq.~\ref{eq:subsequent_spikes}). This opens up interesting possibilities for using the presented networks in recurrent configurations to process continuous input event streams. The backpropagation scheme we outlined in section~\ref{sec:training} would then be analogous to the backpropagation through time (BPTT) technique~\cite{Mozer89} used to train artificial recurrent network.

The presented networks enable very rapid classification of input patterns. As shown in Fig.~\ref{fig:mnist}, the network selects a class before the majority of hidden layer neurons have spiked. This is expected as the only way the network can implement non-linear transformations (in the z-domain) is by changing the causal set of input spikes for each neuron, i.e, by making a neuron spike before a subset of its input neurons have spiked. This unique form of non-linearity not only results in rapid processing, but it enables the efficient implementation of these networks on neuromorphic hardware since processing can stop as soon as an output spike is generated. In the 784-800-10 MNIST network for example, the network classifies an input after only 25 spikes from the hidden layers (on average). Thus, only a small fraction of hidden layer spikes need to be dispatched and processed.

\bibliographystyle{unsrt}
%\bibliography{biblio}

\begin{thebibliography}{10}

\bibitem{LeCun_etal15}
Yann LeCun, Yoshua Bengio, and Geoffrey Hinton.
\newblock Deep learning.
\newblock {\em Nature}, 521(7553):436--444, 2015.

\bibitem{Rumelhart_McClelland86}
D.E. Rumelhart and J.L. McClelland.
\newblock {\em Parallel distributed processing: explorations in the
  microstructure of cognition. Volume 1. Foundations}.
\newblock MIT Press, Cambridge, MA, USA, 1986.

\bibitem{Minsky61}
Marvin Minsky.
\newblock Steps toward artificial intelligence.
\newblock {\em Proceedings of the IRE}, 49(1):8--30, 1961.

\bibitem{Ba_Caruna14}
Jimmy Ba and Rich Caruana.
\newblock Do deep nets really need to be deep?
\newblock In {\em Advances in neural information processing systems}, pages
  2654--2662, 2014.

\bibitem{Memmesheimer_etal14}
Raoul-Martin Memmesheimer, Ran Rubin, B.P. {\"O}lveczky, and Haim Sompolinsky.
\newblock Learning precisely timed spikes.
\newblock {\em Neuron}, 82(4):925--938, 2014.

\bibitem{Gutig_Sompolinsky06}
R.~G{\"u}tig and H.~Sompolinsky.
\newblock The tempotron: a neuron that learns spike timing--based decisions.
\newblock {\em Nature Neuroscience}, 9:420--428, 2006.

\bibitem{Pfister_etal06}
J.-P. Pfister, T.~Toyoizumi, D.~Barber, and W.~Gerstner.
\newblock Optimal spike-timing dependent plasticity for precise action
  potential firing in supervised learning.
\newblock {\em Neural Computation}, 18:1309--1339, 2006.

\bibitem{Gardner_etal15}
Brian Gardner, Ioana Sporea, and Andr{\'e} Gr{\"u}ning.
\newblock Learning spatiotemporally encoded pattern transformations in
  structured spiking neural networks.
\newblock {\em Neural computation}, 2015.

\bibitem{Bohte_etal02}
S.M. Bohte, J.N. Kok, and Han La-Poutre.
\newblock Error-backpropagation in temporally encoded networks of spiking
  neurons.
\newblock {\em Neurocomputing}, 48(1):17--37, 2002.

\bibitem{OConnor_etal13}
P.~O'Connor, D.~Neil, S.-C. Liu, T.~Delbruck, and M.~Pfeiffer.
\newblock Real-time classification and sensor fusion with a spiking deep belief
  network.
\newblock {\em Frontiers in Neuroscience}, 7(178), 2013.

\bibitem{Diehl_etal15}
Peter~U Diehl, Daniel Neil, Jonathan Binas, Matthew Cook, Shih-Chii Liu, and
  Michael Pfeiffer.
\newblock Fast-classifying, high-accuracy spiking deep networks through weight
  and threshold balancing.
\newblock In {\em International Joint Conference on Neural Networks {(IJCNN)}},
  2015.

\bibitem{Cao_etal15}
Yongqiang Cao, Yang Chen, and Deepak Khosla.
\newblock Spiking deep convolutional neural networks for energy-efficient
  object recognition.
\newblock {\em International Journal of Computer Vision}, 113(1):54--66, 2015.

\bibitem{Qiao_etal15}
Ning Qiao, Hesham Mostafa, Federico Corradi, Marc Osswald, Fabio Stefanini,
  Dora Sumislawska, and Giacomo Indiveri.
\newblock A re-configurable on-line learning spiking neuromorphic processor
  comprising 256 neurons and 128k synapses.
\newblock {\em Frontiers in Neuroscience}, 9(141), 2015.

\bibitem{Neil_Liu14}
D.~Neil and S.-C. Liu.
\newblock Minitaur, an event-driven {FPGA}-based spiking network accelerator.
\newblock {\em Very Large Scale Integration ({VLSI}) Systems, {IEEE}
  Transactions on}, PP(99):1--1, October 2014.

\bibitem{Benjamin_etal14}
Ben~Varkey Benjamin, Peiran Gao, Emmett McQuinn, Swadesh Choudhary, Anand~R
  Chandrasekaran, J~Bussat, R~Alvarez-Icaza, JV~Arthur, PA~Merolla, and
  K~Boahen.
\newblock Neurogrid: A mixed-analog-digital multichip system for large-scale
  neural simulations.
\newblock {\em Proceedings of the {IEEE}}, 102(5):699--716, 2014.

\bibitem{Nair_Hinton10}
Vinod Nair and G.E. Hinton.
\newblock Rectified linear units improve restricted {Boltzmann} machines.
\newblock In {\em Proceedings of the 27th International Conference on Machine
  Learning (ICML-10)}, pages 807--814, 2010.

\bibitem{Le-Cun_etal98}
Y.~{Le Cun}, L.~Bottou, Y.~Bengio, and P.~Haffner.
\newblock Gradient-based learning applied to document recognition.
\newblock {\em Proceedings of the {IEEE}}, 86(11):2278--2324, Nov 1998.

\bibitem{Bastien_etal12}
Fr{\'e}d{\'e}ric Bastien, Pascal Lamblin, Razvan Pascanu, James Bergstra, Ian
  Goodfellow, Arnaud Bergeron, Nicolas Bouchard, David Warde-Farley, and Yoshua
  Bengio.
\newblock Theano: new features and speed improvements.
\newblock {\em arXiv preprint arXiv:1211.5590}, 2012.

\bibitem{Bergstra_etal10}
James Bergstra, Olivier Breuleux, Fr{\'e}d{\'e}ric Bastien, Pascal Lamblin,
  Razvan Pascanu, Guillaume Desjardins, Joseph Turian, David Warde-Farley, and
  Yoshua Bengio.
\newblock Theano: a {CPU} and {GPU} math expression compiler.
\newblock In {\em Proceedings of the Python for scientific computing conference
  (SciPy)}, volume~4, page~3. Austin, TX, 2010.

\bibitem{Srivastava_etal14}
Nitish Srivastava, Geoffrey Hinton, Alex Krizhevsky, Ilya Sutskever, and Ruslan
  Salakhutdinov.
\newblock Dropout: A simple way to prevent neural networks from overfitting.
\newblock {\em The Journal of Machine Learning Research}, 15(1):1929--1958,
  2014.

\bibitem{Goodfellow_etal13}
I.J. Goodfellow, David Warde-Farley, Mehdi Mirza, A.C. Courville, and Yoshua
  Bengio.
\newblock Maxout networks.
\newblock {\em In ICML}, 28:1319--1327, 2013.

\bibitem{He_etal16}
Kaiming He, Xiangyu Zhang, Shaoqing Ren, and Jian Sun.
\newblock Deep residual learning for image recognition.
\newblock In {\em Proceedings of the IEEE Conference on Computer Vision and
  Pattern Recognition}, pages 770--778, 2016.

\bibitem{Huang_etal16}
Gao Huang, Zhuang Liu, K.Q. Weinberger, and Laurens van~der Maaten.
\newblock Densely connected convolutional networks.
\newblock {\em arXiv preprint arXiv:1608.06993}, 2016.

\bibitem{Martens10}
James Martens.
\newblock Deep learning via hessian-free optimization.
\newblock In {\em Proceedings of the 27th International Conference on Machine
  Learning (ICML-10)}, pages 735--742, 2010.

\bibitem{Kingma_Ba14}
Diederik Kingma and Jimmy Ba.
\newblock Adam: A method for stochastic optimization.
\newblock {\em arXiv preprint arXiv:1412.6980}, 2014.

\bibitem{Sutskever_etal13}
Ilya Sutskever, James Martens, G.E. Dahl, and G.E. Hinton.
\newblock On the importance of initialization and momentum in deep learning.
\newblock {\em ICML (3)}, 28:1139--1147, 2013.

\bibitem{Huang_etal05}
C.P. Hung, Gabriel Kreiman, Tomaso Poggio, and J.J. DiCarlo.
\newblock Fast readout of object identity from macaque inferior temporal
  cortex.
\newblock {\em Science}, 310(5749):863--866, 2005.

\bibitem{Fabre_etal98}
Mich{\`e}le Fabre-Thorpe, Ghislaine Richard, and S.J. Thorpe.
\newblock Rapid categorization of natural images by rhesus monkeys.
\newblock {\em Neuroreport}, 9(2):303--308, 1998.

\bibitem{Thorpe_etal01}
S.~Thorpe, A.~Delorme, R.~Van Rullen, et~al.
\newblock Spike-based strategies for rapid processing.
\newblock {\em Neural networks}, 14(6-7):715--725, 2001.

\bibitem{Lee_etal16}
Jun~Haeng Lee, Tobi Delbruck, and Michael Pfeiffer.
\newblock Training deep spiking neural networks using backpropagation.
\newblock {\em Frontiers in Neuroscience}, 10, 2016.

\bibitem{Mozer89}
M.C. Mozer.
\newblock A focused back-propagation algorithm for temporal pattern
  recognition.
\newblock {\em Complex systems}, 3(4):349--381, 1989.

\end{thebibliography}


%% \begin{IEEEbiography}{Hesham Mostafa}
%% Biography text here.
%% \end{IEEEbiography}


%\vfill


\end{document}